\documentclass[conference]{IEEEtran}
\IEEEoverridecommandlockouts
\usepackage{cite}
\usepackage{amsmath,amssymb,amsfonts}
\usepackage{algorithmic}
\usepackage{graphicx}
\usepackage{textcomp}
\usepackage{xcolor}
\def\BibTeX{{\rm B\kern-.05em{\sc i\kern-.025em b}\kern-.08em
    T\kern-.1667em\lower.7ex\hbox{E}\kern-.125emX}}

\usepackage{hyperref}
\hypersetup{
     colorlinks = true,
     linkcolor = red,
     anchorcolor = black,
     citecolor = green,
     filecolor = cyan,
     menucolor = red,
     runcolor = cyan, 
     urlcolor = magenta,
     }

\usepackage[nameinlink,noabbrev]{cleveref}

\Crefname{equation}{}{Eqs.}         
\Crefname{figure}{Fig.}{Figs.}

\usepackage{tikz}

\usetikzlibrary{arrows}
\usetikzlibrary{positioning, shapes.geometric, shapes.misc}

\usepackage{multirow}
\usepackage{siunitx}
\usepackage{bbm}
\usepackage[normalem]{ulem}

\newcommand{\IoU}{\mathit{IoU}}

\newcommand{\sIoU}{\mathit{IoU}_{\mathrm{adj}}}

\DeclareMathOperator*{\argmax}{arg\,max}

\newcommand{\intr}{{in}}
\newcommand{\bdr}{{bd}}
\newcommand{\nb}{{nb}}

\newcommand{\SP}{\mathit{SP}}

\newcommand{\figwidth}{0.48}
\newcommand{\tabscale}{0.68}

\begin{document}

\title{False Positive Detection and Prediction Quality Estimation for LiDAR Point Cloud Segmentation
\thanks{H.G. and M.R. acknowledge financial support through the research consortium bergisch.smart.mobility funded by the ministry for economy, innovation, digitalization and energy (MWIDE) of the state North Rhine Westphalia under the grant-no. DMR-1-2.}
}

\author{\IEEEauthorblockN{1\textsuperscript{st} Pascal Colling}
\IEEEauthorblockA{\textit{Department of Mathematics} \\
\textit{University of Wuppertal, Germany}\\
pascal.colling@uni-wuppertal.de}
\and
\IEEEauthorblockN{2\textsuperscript{nd} Matthias Rottmann}
\IEEEauthorblockA{\textit{Department of Mathematics} \\
\textit{University of Wuppertal, Germany}\\
rottmann@uni-wuppertal.de} \\
\IEEEauthorblockN{4\textsuperscript{th} Hanno Gottschalk}
\IEEEauthorblockA{\textit{Department of Mathematics} \\
\textit{University of Wuppertal, Germany}\\
hanno.gottschalk@uni-wuppertal.de}
\and
\IEEEauthorblockN{3\textsuperscript{rd} Lutz Roese-Koerner}
\IEEEauthorblockA{\textit{Aptiv, Wuppertal, Germany} \\
lutz.roese-koerner@aptiv.de}
}

\maketitle
 
\begin{abstract}
We present a novel post-processing tool for semantic segmentation of LiDAR point cloud data, called LidarMetaSeg, which estimates the prediction quality segmentwise. For this purpose we compute dispersion measures based on network probability outputs as well as feature measures based on point cloud input features and aggregate them on segment level. These aggregated measures are used to train a meta classification model to predict whether a predicted segment is a false positive or not and a meta regression model to predict the segmentwise intersection over union. Both models can then be applied to semantic segmentation inferences without knowing the ground truth.
In our experiments we use different LiDAR segmentation models and datasets and analyze the power of our method. We show that our results outperform other standard approaches.
\end{abstract}


\section{Introduction} \label{sec:introduction}

\begin{figure}[ht]
    \centering
    \centerline{\includegraphics[width=0.45 \textwidth]{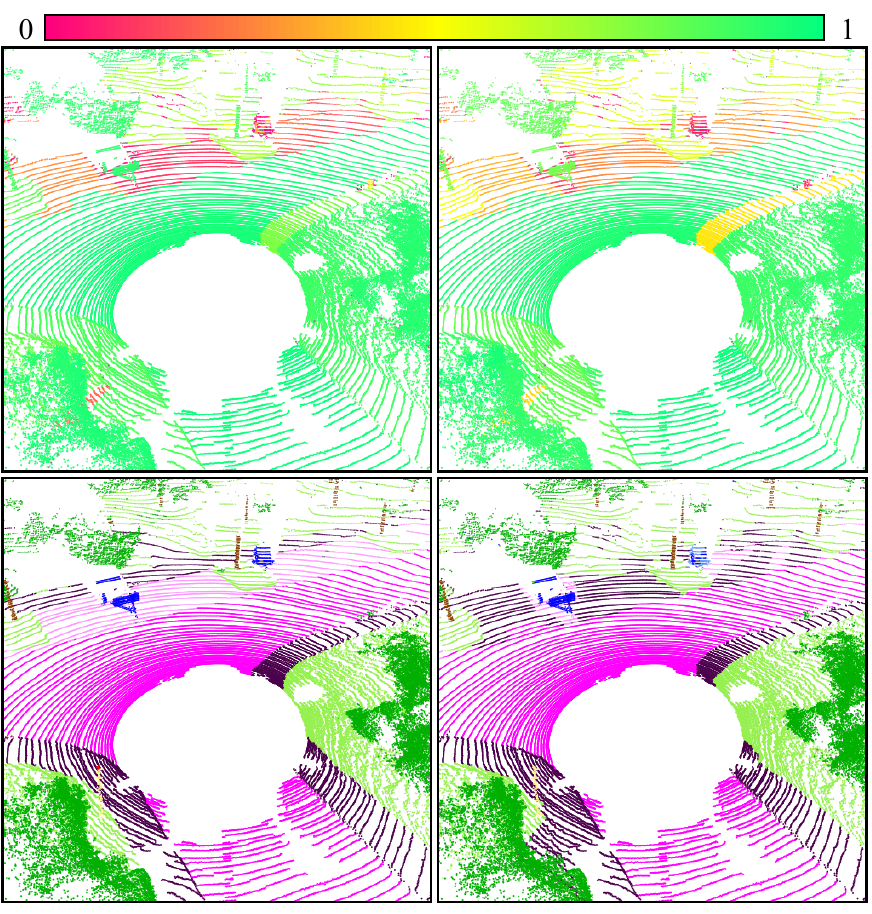}}
    \caption{A visualization of LidarMetaSeg containing the ground truth (bottom left), the LiDAR segmentation (bottom right), the LiDAR segmentation quality (top left) as the $\IoU$ of prediction and ground truth and its estimation obtained by LidarMetaSeg (top right).
    The higher the $\IoU$, the better the prediction quality.}
    \label{fig:01_LidarMetaSeg}
\end{figure}

In the field of automated driving, scene understanding is essential. One possible solution for the semantic interpretation of scenes captured by multiple sensor modalities is LiDAR point cloud segmentation \cite{milioto2019rangenet++, cortinhal2020salsanext, zhou2020cylinder3d, xu2021rpvnet} (in the following LiDAR segmentation for brevity) where each point of the point cloud is assigned to a class of a given set. A segment is an area of points of the same class.
Compared to camera images, a LiDAR point cloud is relatively sparse, but provides accurate depth information. Furthermore, since the LiDAR sensor in general is rotating, 360 degrees of the environment are considered. A summary for sensor modalities is given in \cite{wang2019multi}.
In recent years, the performance of LiDAR segmentation networks has increased enormously \cite{qi2017pointnet, milioto2019rangenet++, cortinhal2020salsanext, zhou2020cylinder3d, xu2021rpvnet}, but there are only few works on uncertainty quantification \cite{cortinhal2020salsanext}. In applications of street scene understanding, safety and reliability of perception systems are just as important as their accuracy.
To tackle this problem, we introduce a post-processing tool, called \emph{LidarMetaSeg}, which estimates the segmentwise (i.e., per connected component of the predicted segmentation) prediction quality in terms of segmentwise intersection over union \cite{jaccard1912distribution} ($\IoU$) of the LiDAR segmentation model, see also \cref{fig:01_LidarMetaSeg}. This provides not only uncertainty quantification per predicted segment but also an online assessment of prediction quality.

State-of-the-art LiDAR segmentation models are based on deep neural networks and can be grouped into two main approaches: projection-based (2D) and non-projection-based (3D) networks, cf. \cite{guo2020deep_lidar_srv}.
Projection-based networks like \cite{xu2020squeezesegv3, milioto2019rangenet++, cortinhal2020salsanext} use a spherical (2D) image representation of the point cloud. The predicted semantic categories on the image are thereafter reinserted along the spherical rays into the 3D point cloud. This may contain some post-processing steps, like the k-nearest neighbor (kNN) approach, see \cite{milioto2019rangenet++}. Due to the representation of point clouds as projected images, the networks employed for LiDAR segmentation have architectures that often resemble image segmentation architectures. 
The non-projection-based networks, e.g.\ \cite{qi2017pointnet++, thomas2019kpconv, zhou2020cylinder3d}, process the point cloud directly in 3D space with or without different 3D representation approaches. For example, in \cite{thomas2019kpconv}, the network operates on the 3D point cloud without introducing an additional representation while in \cite{zhou2020cylinder3d} the authors perform a 3D cylinder partition. A combination of a 2D and 3D representation of the point cloud is used in \cite{xu2021rpvnet}.
All current architectures, using a 2D or 3D representation or a combination of both provide the segmentation of the point cloud. Therefore, it is also possible to output the probabilities, which is the only prerequisite required for LidarMetaSeg. 

Concerning uncertainty quantification in deep learning, Bayesian approaches like Monte Carlo (MC) dropout \cite{gal2016dropout} are commonly used, e.g.\ in image-based object detection \cite{ozdemir2017propagating_image_mc_det}, image segmentation \cite{kendall2015bayesian_image_mc_seg} and also in LiDAR object detection \cite{feng2018towards_lidar_mc_det}. In object detection and instance segmentation, so called scores containing (un)certainty information are used, while this is not the case for semantic segmentation.
The network SalsaNext \cite{cortinhal2020salsanext} is for LiDAR segmentation and makes use of MC dropout to output the model (epistemic) and observation (aleatoric) uncertainty. 

In our method LidarMetaSeg we first project the point cloud and the corresponding softmax probabilities of the network to a spherical 2D image representation, which are then used to compute different types of dispersion measures resulting in different dispersion heatmaps.
To estimate uncertainty on segment level, we aggregate the dispersion measures with respect to each predicted segment.
The $\IoU$ is commonly used to evaluate the performance of a segmentation model.
For each predicted segment, we compute its $\IoU$ with the ground truth and call this \emph{segmentwise $\IoU$}. In our experiments we observe a strong correlation of the segmentwise $\IoU$ with the aggregated dispersion measures.
Hence, we use the aggregated dispersion measures with additional information from the point cloud input to create a set of handcrafted features. The latter are used in post-processing manner as input for training \emph{i)} a \emph{meta classification model} to detect false positive segments, i.e., if the $\IoU$ is equal or greater than $0$ and \emph{ii)} a \emph{meta regression model} to estimate the segmentwise $\IoU$.
Thus, we not only have a pointwise uncertainty quantification, given by the dispersion heatmaps, but also a false positive detection as well as a segmentation quality estimation on segment level. 

The idea of meta classification and regression to detect false positives and to estimate the segmentwise prediction quality was first introduced in the field of semantic segmentation of images \cite{rottmann2020prediction}, called MetaSeg. The work presented in \cite{roy2018inherent} goes in a similar direction, but for brain tumor segmentation. 
MetaSeg was further extended in other directions, i.e., for controlled false negative reduction \cite{Chan2019metafusion}, for time dynamic uncertainty estimates for video data \cite{Maag2019}, for taking resolution-dependent uncertainties into account \cite{Schubert2019} and as part of an active learning method \cite{colling2021metabox}.
Inspired by the possibility of representing the point could as a 2D image, our method LidarMetaSeg is an extension and further development of the original work.
Therefore MetaSeg \cite{rottmann2020prediction} is the most related work to our approach LidarMetaSeg, which up to now together with SalsaNext \cite{cortinhal2020salsanext} are the only works in the direction of uncertainty quantification in LiDAR segmentation. 

With MC dropout, SalsaNext follows a Bayesian approach to quantifying the model and the observation uncertainty. The uncertainty output is point-based and not segment-based, as in our approach.
Also for MC dropout, the model has to infer one sample multiple times. 
LidarMetaSeg requires only a single network inference and estimates uncertainties by means of the network's class probabilities. In a 2D representation, these pixelwise uncertainty estimates can be viewed as uncertainty heatmaps. From those heatmaps, we compute aggregated uncertainties for each predicted segment, therefore clearly going beyond the stage of pixelwise uncertainty estimation. In contrast to MetaSeg for image segmentation, we not only use the network's output but also utilize information from the point cloud input, such as the intensity and range features provided for each point of the point cloud.

LidarMetaSeg is therefore a universal post-processing tool that allows for the detection of false positive segments as well as the estimation of segmentwise LiDAR segmentation quality. Besides that, the present work is the first one to provide uncertainty estimates on the level of predicted segments.
We evaluate our method on two different datasets, SemanticKITTI \cite{behley2019semantickitti} and nuScenes \cite{caesar2020nuscenes} and with three different network architectures, two projection-based models RangeNet++ \cite{milioto2019rangenet++}, SalsaNext \cite{cortinhal2020salsanext} and one non-projection-based model, Cylinder3D \cite{zhou2020cylinder3d}. For meta classification, we achieve area under receiver operating characteristic curve (AUROC) and area under precision recall curve (AUPRC) \cite{davis2006relationship_auroc} values of up to $91.16 \% $ and $74.35 \%$, respectively. For the meta regression, we achieve coefficient of determination $R^2$ values of up to $66.69 \%$.
We show that our aggregated measures -- in terms of meta classification and regression -- lead to a significant performance gain in comparison to when only considering a single uncertainty metric like the segmentwise entropy.

\section{Method} \label{sec:method}

LidarMetaSeg is a post-processing method for LiDAR semantic segmentation to estimate the segmentwise prediction quality. It consists of a meta classification and a meta regression model that for each predicted segment classifies whether it has an $\IoU$ equal to or greater than $0$ with the ground truth and predicts the segmentwise $\IoU$ with the ground truth, respectively.
The method works as follows: in a preprocessing step we project each sample, i.e., the point cloud, the corresponding network probabilities and the labels into a spherical 2D image representation. In a next step and based on the projected data, we compute dispersion measures and other features like it is done for image data in \cite{rottmann2020prediction, Schubert2019, Chan2019metafusion}.
Afterwards we identify the segments of a given semantic class and aggregate the pixelwise values from the previous step on a segment level. In addition, we compute the $\IoU$ of each predicted segment with the ground truth of the same class. This results in a structured dataset, which consist of the coefficients of the aggregated dispersion measures as well as additional features and of the target variable -- the $\IoU \in [0, 1] $ for the task of meta regression or the binary variable $IoU=0, >0$ ($IoU=0$ as indicator for a false positive) for the task of meta classification -- for each segment. We fit a classification and a regression model to this dataset. In the end, we re-project the meta classification and regression from the image representation to the point cloud.

\subsection{Preprocessing} \label{subsec:preprocessing}

A sample of input data for LidarMetaSeg is assumed to be given on point cloud level and contains the following:
\begin{itemize}
    \item \emph{point cloud} $\tilde{p} \in \mathbb{R}^{m \times 4}, \ \tilde{p}_j = (x_j, y_j, z_j, i_j)$ with $x_j, y_j, z_j \in \mathbb{R}$ Cartesian coordinates and intensity $i_j\in \mathbb{R}_+$ for $j=1,\ldots,m$ with $m$ the number of points in the LiDAR point cloud,
    
    \item \emph{ground truth / labels} $\tilde{y}^* \in\mathcal{C}^m$ with $\mathcal{C} = \{ 1, \dots, n\}$ the set of $n$ given classes,
    
    \item \emph{probabilities} $\tilde{y}^{\mathit{prob}} = f(\tilde{p}) \in \mathbb{R}_{ [0, 1] }^{m \times n}$ of a LiDAR segmentation network $f$, given as softmax probabilities,
    
    \item \emph{prediction} $\tilde{y} = \underset{c \in \mathcal{C}}{\argmax} \ \tilde{y}^{\mathit{prob}}$ \text{.}
\end{itemize}
Typically, one is also interested in the range of a given point in the point cloud, which is part of most LiDAR segmentation networks' input. Since the ego car is located in the origin of the coordinate system, this quantity is given by $r_j = \sqrt{x_j^2 + y_j^2 + z_j^2}$ for each $\tilde{p}_j$.

The projection of a point cloud to a spherical 2D image representation follows two steps: a transformation from Cartesian to spherical coordinates and then a transformation from spherical to image coordinates.
The spherical coordinates are given as $(r_j, \theta_j, \varphi_j)$ with range $r_j$, polar angle $\theta_j$ and azimuth angle $\varphi_j$. The transformation for the Cartesian to spherical coordinates is given by
\begin{equation}
    \theta_j = \arcsin 	\left( \frac{z_j}{r_j} \right) \quad \text{and} \quad
    \varphi_j = \arctan \left( \frac{y_j}{x_j} \right)
\end{equation}
and $r_j$ for $j=1,\ldots,m$. Based on the spherical coordinates we get the image coordinates $(u,v)$ with the equation

\begin{align}
    \begin{pmatrix}
        u \\
        v
    \end{pmatrix}
    = 
    \begin{pmatrix}
        \frac{1}{2} \bigl\lbrack  1 - ( \varphi_j \pi^{-1} ) \bigr\rbrack w \\
        \bigl\lbrack 1 - ( \theta_j + f_{\mathit{ver}^+}  ) f_{\mathit{ver}}^{-1}  \bigr\rbrack h
    \end{pmatrix}
\end{align}
with $(w, h)$ the width and height of the image and $f_{\mathit{ver}} = f_{\mathit{ver}^+} + f_{\mathit{ver}^-} $ the vertical field of view (FOV) of the LiDAR sensor. 

In order to get an image representation where each point correspond to one pixel and vice versa, we need the number of channels, the angular resolution and the horizontal FOV of the LiDAR sensor.
To this end, we define the height $h$ as the number of channels and the width $w$ as the quotient of the horizontal FOV $f_{\mathit{hor}}$ (which is in general $360$, since the LiDAR sensor is rotating) and the angular resolution $\alpha$, i.e., 
\begin{equation} \label{eq:w}
     w = \frac{f_{\mathit{hor}}}{\alpha} \text{.}
\end{equation}
Thus, the image representation -- using the explicit sensor information -- has as many entries or pixels as the point cloud can have as maximum number of points. Unfortunately there are still some technical reasons, due to which it can happen that multiple points are projected to the same pixel, e.g.\ ego-motion compensation or overlapping channel angles. More details concerning such projection errors can be found in \cite{triess2020scan}. With the projection proposed above this happens rarely enough, so that this event is negligible.

\begin{figure}
    \centering
    \centerline{\includegraphics[width=\figwidth\textwidth]{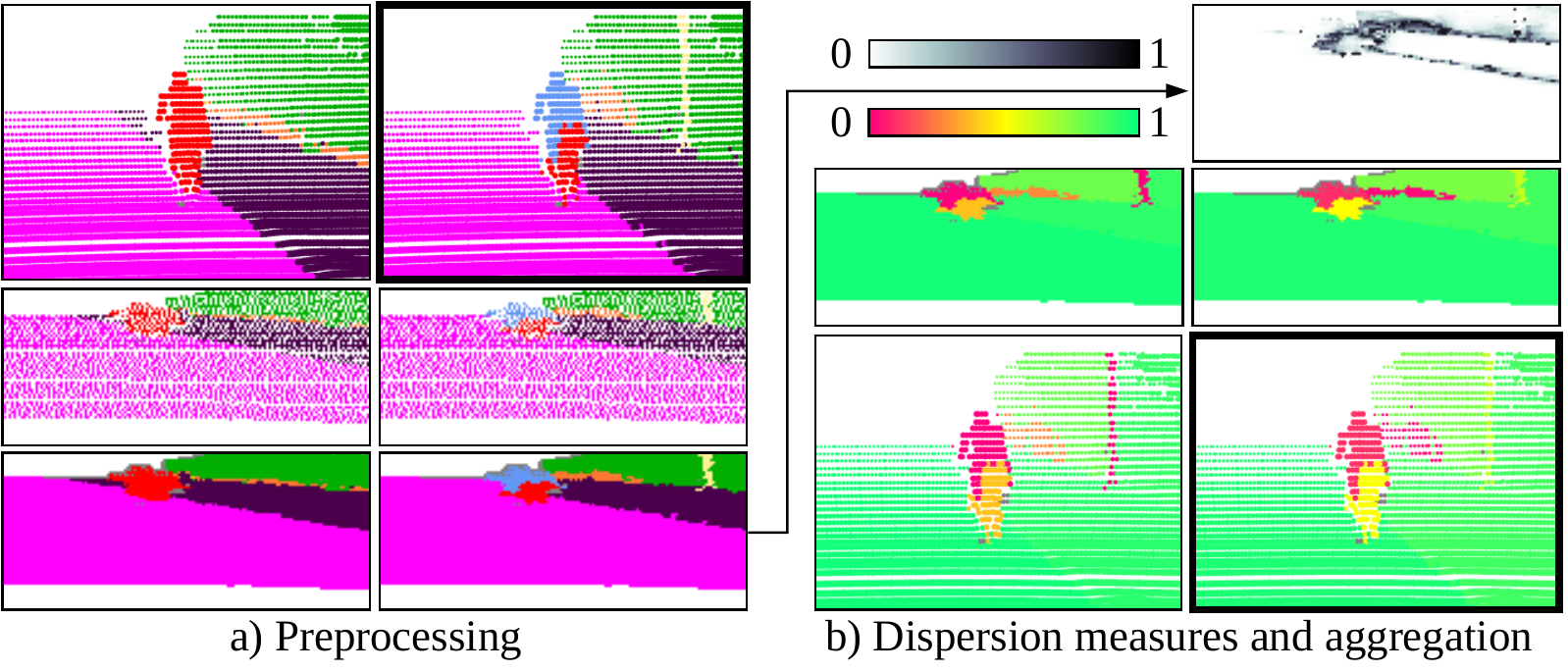}}
    \caption{
    Visual examples of our method LidarMetaSeg. The left panel shows the preprocessing part: the ground truth (top left) and the prediction (top right) of the point cloud as well as the corresponding sparse (middle) and filled (bottom) image representations.
    The right panel visualizes a dispersion heatmap, the segmentwise prediction quality and its estimation: the probability difference heatmap of the prediction-based probabilities (top right), where higher values correspond to higher uncertainty, in the middle the true (left) and estimated (right) $\sIoU$ values for the image representation and in the bottom part the corresponding visualizations after the re-projection to the point cloud. The prediction of the point cloud and the corresponding prediction quality estimation is highlighted.}
    \label{fig:02_method}
\end{figure}

Following the projection above, we denote the projected 2D representation similar as before, but without $\tilde{\cdot}$, i.e.,

\begin{itemize}
    \item image representation (of the point cloud $\tilde{p}$) \\ $F = ( F^x, F^y, F^z, F^i, F^r ), \ F \in \mathbb{R}^{w \times h \times 5}$,
    \item ground truth / labels $y^* \in \mathcal{C}^{w \times h} $,
    \item probabilities $y^{\mathit{prob}} \in \mathcal{C}^{w \times h \times n} $,
    \item prediction $y \in \mathcal{C}^{w \times h} $.
\end{itemize}
The proposed image projection yields a sparse image representation. 
However, our post-processing approach LidarMetaSeg is based on connected components of the segmentation. 
In order to identify connected components (segments) of pixels in the 2D image resulting from the projection, we fill these gaps by setting any empty entry $l := (u, v)$ (entries without a corresponding point in the point cloud) to a value of one of its nearest neighbors that received a value from the projection.
An example of such a filled image representation is shown in \cref{fig:02_method}, left panel.
In the following we only consider the filled image representations. 
We store the information which pixel received its value via projection and which one via fill-in in a binary mask of width $w$ and height $h$ denoted by $\delta$ where $1$ represents a projected point and $0$ a filled entry, i.e., 
\begin{align}
\delta_l = \begin{cases}
1, & l \text{ corresponds to a projected point,} \\
0, &   \text{else.}
\end{cases}
\end{align}
For simplicity, we refer the filled image representations $F$ (that are input quantities for the segmentation network) as feature measures.

\subsection{Dispersion Measures and Segmentwise Aggregation} \label{subsec:comp_metrics}

First we define the dispersion and feature measures and afterwards the segmentwise aggregation.
\paragraph{Dispersion and Feature Measures}
Based on the probabilities $y^{\mathit{prob}} \in [0,1]^{w \times h \times n}$, we define the dispersion measures \emph{entropy} $E_l$, \emph{probability difference} $D_l$ and \emph{variation ratio} $V_l$ at pixel position $l=(u,v)$ as follows:
\begin{align}
    & E_l = - \frac{1}{ \log (n) } \sum_{c=1}^{n} y^{\mathit{prob}}_{l,c} \ \log \bigl( y^{\mathit{prob}}_{l,c} \bigr) \text{,} \label{eq: disp_e} \\
    & D_l = 1 - \max_{c_1 \in \mathcal{C}} y^{\mathit{prob}}_{l,c_1} + \max_{c_2 \in \mathcal{C} \setminus c_1} y^{\mathit{prob}}_{l,c_2} \text{,} \label{eq: disp_d} \\
    & V_l = 1 - \max_{c \in \mathcal{C}} y^{\mathit{prob}}_{l,c} \label{eq: disp_v} \text{.}
\end{align}
In addition, the feature measures \emph{coordinates}, \emph{intensity} and \emph{range} at position $l$ are given by the image representation
\begin{align}
    F_l^{\sharp}, \ \sharp \in \{ x, y, z, i, r\} \text{.} 
\end{align} 
For the sake of brevity, we define the set of dispersion and features measures 
\begin{align}
    \mathcal{M} = \{ E, D, V, F^x, F^y, F^z, F^i, F^r \}
\end{align}
omitting the index for the position $l$ as this will follow from the context. Note that, due to the position dependence, each element of $\mathcal{M}$ can be considered as a heatmap.

\paragraph{Segmentwise Aggregation}
For a given prediction $y \in \mathcal{C}^{ w \times h }, \ \mathcal{C} = \{ 1, \dots , n\}$ and the corresponding ground truth $y^* \in \mathcal{C}^{ w \times h }, \ \mathcal{C} = \{ 1, \dots , n\}$, we denote $\mathcal{K}_y$ and $\mathcal{K}_{y^*}$ the set of connected components (segments) in the prediction and the ground truth, respectively.
A connected component $k$ is a set of pixels that are adjacent to each other and belong to the same class, see also \cref{fig:02_method}, left panel.
For each segment $k \in \mathcal{K}_{y}$, we define the following quantities. Additionally and in order to count only the pixels with a corresponding point in the point cloud, we introduce the restriction by the corresponding binary mask $\delta$ with $|\cdot|_{\delta}$.

\begin{itemize}
    \setlength\itemsep{-0.15em}
    \item The interior $k_\intr = \{ (u,v) \in k: [ u \pm 1 ] \times [v \pm 1 ] \in k\} \subset k$, i.e., a pixel $l = (u,v)$ is an element of $k_\intr$ if all eight neighboring pixels are an element of $k$,
    \item the boundary $k_\bdr= k \setminus k_\intr$,
    \item the pixel sizes $S=|k|$, $S_\intr=|k_\intr|$, $S_\bdr=|k_\bdr|$,
    \item the segment size in the point cloud $\SP= |k|_{\delta}$.
 \end{itemize}
Furthermore, we define the target variables $\IoU$ and the so called adjusted $\sIoU$ as follow:
 \begin{itemize}
    \item $\IoU$: let ${\mathcal K}_{y^*}|_k$ be the set of all $k'\in \mathcal{K}_{y^*}$ that have non-trivial intersection with $k$ and whose class label equals the predicted class for $k$, then
    \begin{align}
    \IoU(k) = \frac{|k \cap K'|_{\delta}}{|k \cup K'|_{\delta}}\,,\qquad K' = \bigcup_{k' \in {\mathcal K}_{y^*} |_k} k' \text{,}
    \end{align}
    \item the adjusted $\sIoU$ does not count pixels in the ground truth segment that are not contained in the predicted segments, but in other predicted segments of the same class: let $Q = \{ q \in \mathcal{K}_y: q \cap K' \neq \emptyset \} $, then
    \begin{align}
    \sIoU(k) = \frac{|k \cap K'|_{\delta}}{|k \cup (K' \setminus Q)|_{\delta}} \text{.} 
    \end{align}
 \end{itemize}
In cases where a ground truth segment is covered by more than one predicted segment of the same class, each predicted segment would have a low $\IoU$, while the predicted segments represent the ground truth quite well. As a remedy, the adjusted $\sIoU$ was introduced in \cite{rottmann2020prediction} to not punish this situation. The adjusted $\sIoU$ is more suitable for the task of meta regression.
For the meta classification it holds $\IoU =0, > 0 \Leftrightarrow \sIoU =0, > 0 $.

Based on the previous definitions, we define the dispersion and feature measures:
 \begin{itemize}
    \item the mean $\mu$ and variance $\upsilon$ metrics
    \begin{align}
        &\mu M_{\sharp} := \mu ( M_{\sharp} ) = \frac{1}{S_{\sharp}} \sum_{l \in k_{\sharp}} M_l  \\
        &\upsilon M_{\sharp} := \upsilon (M_{\sharp})= \mu ( M_{\sharp}^2) - \mu (M_{\sharp} )^2
    \end{align}
    for $\sharp \in \{\_,in,bd\}$ and $ M \in \mathcal{M}$,
    \item the relative sizes $\bar S = S/S_\bdr$, $\bar S_\intr = S_\intr/S_\bdr$,
    \item the relative mean and variance metrics
    \begin{align}
        &\tau \bar M = \tau M \bar S \label{eq:mean_rel_bd} \\
        &\tau \bar M_\intr = \tau M \bar S_\intr \label{eq:mean_rel_in}
    \end{align}
    for $\tau \in \{ \mu, \upsilon\}$ and $M \in \mathcal{M}$,
    \item the ratio of the neighborhood's correct predictions of each class
    \begin{align}
        N_c = \frac{1}{| k_\nb| } \sum_{l \in k_\bdr} \mathbbm{1}_{ \{ c = y_l \} } \qquad \forall c \in \mathcal{C} 
    \end{align}
    with $k_\nb$ the set of $k$ neighbors, i.e., \ $k_\nb = \{ l' \in [ u \pm 1 ] \times [v \pm 1] \subset w \times h : (u, v) \in k, l' \notin k\}$,
    \item the mean class probabilities
    \begin{align}
        P_c = \frac{1}{S} \sum_{l \in k } y_{l, c}^{ \mathit{prob} } \qquad \forall c \in \mathcal{C} \text{.}
    \end{align}
\end{itemize}
Typically, the dispersion measures $E_l, D_l, V_l$ are large for $l \in k_\bdr$. This motivates the separate treatment of interior and boundary measures. Furthermore we observe a correlation between fractal segment shapes and a bad or wrong prediction, which motivates the relative sizes $\bar S, \ \bar S_\intr$.
In summary, we have $86 + 2n$ metrics: the (relative) mean and variance metrics $\tau M , \tau M_\intr, \tau M_\bdr, \tau \bar{M}, \tau \bar{M}_\intr \ \forall \tau \in \{ \mu, \upsilon\}, \ \forall M \in \mathcal{M}$, the (relative) size metrics $S, S_\intr, S_\bdr, \bar{S}, \bar{S}_\intr, \SP$ as well as $N_c, P_c \ \forall c \in \mathcal{C}.$
An example of the pixelwise dispersion measures as well as the segmentwise $\sIoU$ values and its prediction is shown in \cref{fig:02_method}, right panel.

With the exception of the segmentwise $\IoU$ and $\sIoU$ values, all quantities defined above can be computed without the knowledge of the ground truth.

\section{Numerical Experiments} \label{sec:results}

For numerical experiments we used two datasets: SemanticKITTI \cite{behley2019semantickitti} and nuScenes \cite{caesar2020nuscenes}. For meta classification and regression we deploy XGBoost \cite{chen2016xgboost}. Other classification and regression methods like linear / logistic regression, neural networks of tree based ensemble methods \cite{friedman2001elements} are also possible. However, as shown in \cite{Maag2019}, XGBoost leads to the best results. Due to the reason mentioned in the previous section, the target variable for the meta regression (and classification) is the adjusted $\sIoU$.

First, we describe the settings of the experiments for both datasets and evaluate the results for the false positive detection and the segmentwise prediction quality estimation when using all metrics presented in the previous section. Afterwards we conduct an analysis of the metrics and the meta classification model. 

\subsection{SemanticKITTI} \label{subsec:kitti}

The SemanticKITTI dataset \cite{behley2019semantickitti} contains street scenes from and around Karlsruhe, Germany. It provides $11$ sequences with about $23$K samples for training and validation, consisting of $19$ classes. The data is recorded with a Velodyne HDL-64E LiDAR sensor, which has $64$ channels and a (horizontal) angular resolution of \ang{0.08}. Furthermore the data is recorded and annotated with $10$ frames per second (fps) and each point cloud contains about $120$K points.
The authors of the dataset recommend to use all sequences to train the LiDAR segmentation model, except sequence $08$, which should be used for validation. 

For the experiments we used three pretrained LiDAR segmentation models, two projection-based models, i.e., RangeNet++ \cite{milioto2019rangenet++} and SalsaNext \cite{cortinhal2020salsanext}, and one non-projection-based model, i.e., Cylinder3D \cite{zhou2020cylinder3d}, which followed the recommended data split. 
For RangeNet++ and SalsaNext, the softmax probabilities are given for the  2D image representation prediction. As we assume that softmax probabilities are given for the point cloud, we consider this representation as the starting point and re-project the softmax probabilities to the point cloud.

After the re-projection from the 2D image representation prediction to the point cloud, both models have an additional kNN post-processing step to clean the point cloud from undesired discretization and inference artifacts \cite{milioto2019rangenet++}, which may results in changing the semantic class of a few points.
To take this post-processing step into account, we set the values of the cleaned points in the corresponding softmax probabilities of the point cloud to $1$ and all other values to $0$.
Therefore the softmax condition (the sum of all probability values of a point is equal to $1$ and all values are between $0$ and $1$) is met and the adjusted prediction is equal to the argmax of the probabilities.
We do not expect other approaches to significantly change the results since we aggregate our dispersion measures and the number of modified points is small.

Following our method, the image representation of the point cloud data is of size $(w, h) = (4500, 64)$, cf.\ \cref{eq:w}.

Most deep learning models tend to overfit. Therefore we only use samples for LidarMetaSeg, which are not part of the training data of the segmentation network, as overfitted models affects the dispersion measures.
Thus, we only use sequence $08$ for our experiments.
Computing the connected components and metrics yields approx.\ $3.4$M segments for each network. Most of the segments are very small. Therefore we follow a similar segment exclusion rule as in MetaSeg \cite{rottmann2020prediction}, where segments with empty interior, $S_{\mathit{in}} = 0$, are excluded. Here, we exclude segments consisting of less than $10$ LiDAR points, i.e., $ \SP< 10 $, also shown in gray color in \cref{fig:02_method}. Hence, we reduce the number of segments to approx. $0.45$M but we retain $99 \%$ of the data measured in terms of the number of points. We tested the dependence of our results under variation of the exclusion size $\SP$. The results were very similar to the results we present in the following.

For training and validation of LidarMetaSeg we split sequence $08$ and the corresponding connected components and metrics into $10$ disjoint sub-sequences. These sub-sequences are used for a $10$-fold cross validation. A cross validation over all samples would yield highly correlated training and validation splits as all sequences are recorded with 10 fps.
The results for the meta classification and regression are given in \cref{tab:summary}. 
For all three models we achieve a validation accuracy between $85.50 \%$ and $88.37 \% $, see row `ACC LMS' (short for LidarMetaSeg). The accuracy of random guessing (`ACC naive baseline') is between $ 78.42 \%$ and $84.53 \%$ which directly amounts to percentage of segments with an $\sIoU > 0$.

For each method, the accuracy values correspond to a single decision threshold. In contrast to that, the AUROC and AUPRC are obtained by varying the decision threshold of the classification output.
The AUROC essentially measures the overlap of distributions corresponding to negative and positive samples; this score does not place more emphasis on one class over the other in case of class imbalance. 
The ACC of random guessing indicates the class imbalance: about $80 \%$ of the segments have an $\sIoU >0$ and $20 \%$ of the segments have an $\sIoU = 0$, i.e., they are false positives. The underlying precision recall curve of the AUPRC ignores true negatives and emphasizes the detection of the positive class (false positives).

\begin{figure}
    \centering
    \centerline{\includegraphics[width=\figwidth \textwidth]{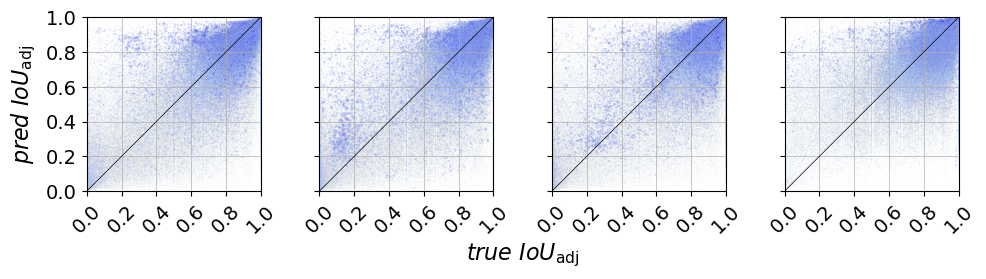}}
    \caption{True $\mathit{IoU}_{\mathit{adj}}$ vs predicted $\mathit{IoU}_{\mathit{adj}}$ for RangeNet++, SalsaNext, Cylinder3D on SemanticKITTI as well as Cylinder3D on nuScenes, from left to right.}
    \label{fig:03_scatter_pred_true_iou}
\end{figure}

Using the metrics of the previous section (LMS) for the meta classification yields AUROC values above $90 \%$ and AUPRC up to $74.35 \%$. 
For the meta regression we achieve $R^2$ values between $61.57 - 66.69 \%$. 
\Cref{fig:03_scatter_pred_true_iou} depicts the quality of predicting the $\sIoU$. A visualization of estimating the $\sIoU$ is shown in \cref{fig:02_method} and in the supplementary video\footnote{\url{https://youtu.be/907jJSRgHUk}}.

\begin{table*}
\begin{center}
\caption{Results for meta classification and regression, averaged over 10 runs. The numbers in the brackets denote standard deviations of the computed mean values. The best results in terms of ACC, AUROC, AUPRC and $R^2$ on the validation data are highlighted. \label{tab:summary}}
\scalebox{\tabscale}{
\begin{tabular}{|l|l|r|r||r|r||r|r||r|r|} \cline{3-10}
\multicolumn{2}{c|}{} & \multicolumn{6}{|c||}{SemanticKITTI} & \multicolumn{2}{|c|}{nuScenes} \\ \cline{3-10}
\multicolumn{2}{c|}{} & \multicolumn{2}{|c||}{RangeNet++} & \multicolumn{2}{|c||}{SalsaNext}  & \multicolumn{2}{|c||}{Cylinder3D} & \multicolumn{2}{|c|}{Cylinder3D} \\ \cline{3-10}
\multicolumn{2}{c|}{} & \multicolumn{1}{|c|}{training} & \multicolumn{1}{|c||}{validation} & \multicolumn{1}{|c|}{training} & \multicolumn{1}{|c||}{validation} & \multicolumn{1}{|c|}{training} & \multicolumn{1}{|c||}{validation} & \multicolumn{1}{c|}{training} & \multicolumn{1}{|c|}{validation}\\
\hline
\multicolumn{10}{|c|}{Classification $\sIoU=0,>0$} \\
\hline
\multirow{5}{*}{\rotatebox{90}{ACC}} & LMS                        & $92.26\%(\pm0.25\%)$ & $\mathbf{85.50\%}(\pm3.12\%)$ & $92.43\%(\pm0.21\%)$ & $\mathbf{86.26\%}(\pm2.75\%)$ & $92.97\%(\pm0.21\%)$ & $\mathbf{88.37\%}(\pm2.89\%)$ & $92.96\%(\pm0.08\%)$ & $\mathbf{91.00\%}(\pm0.60\%)$  \\
& LMS w/o features   & $90.64\%(\pm0.30\%)$ & $85.10\%(\pm3.21\%)$ & $90.77\%(\pm0.23\%)$ & $86.02\%(\pm2.86\%)$ & $91.68\%(\pm0.27\%)$ & $88.22\%(\pm2.93\%)$ & $92.31\%(\pm0.13\%)$ & $90.62\%(\pm1.16\%)$ \\
& Entropy                    & $79.37\%(\pm0.37\%)$ & $79.21\%(\pm3.17\%)$ & $80.14\%(\pm0.33\%)$ & $80.06\%(\pm2.75\%)$ & $85.35\%(\pm0.30\%)$ & $85.24\%(\pm2.78\%)$ & $89.91\%(\pm0.14\%)$ & $89.86\%(\pm1.17\%)$\\
& LMS $\cup$ MCDO    &                      &                      & $92.48\%(\pm0.20\%)$ & $86.26\%(\pm2.78\%)$ &                      &               &  &        \\ \cline{2-10}
& naive baseline        & \multicolumn{2}{|c||}{$78.42\%$} & \multicolumn{2}{|c||}{$79.36\%$} & \multicolumn{2}{|c||}{$84.53\%$}  & \multicolumn{2}{|c|}{$89.85\%$}\\ 
\hline
\multirow{4}{*}{\rotatebox{90}{AUROC}} & LMS                      & $97.19\%(\pm0.12\%)$ & $\mathbf{90.58\%}(\pm1.89\%)$ & $97.22\%(\pm0.10\%)$ & $91.16\%(\pm1.75\%)$ & $96.86\%(\pm0.10\%)$ & $\mathbf{90.78\%}(\pm2.36\%)$ & $94.81\%(\pm0.10\%)$ & $\mathbf{90.05\%}(\pm1.11\%)$  \\
& LMS w/o features & $95.97\%(\pm0.13\%)$ & $89.85\%(\pm1.96\%)$ & $95.95\%(\pm0.10\%)$ & $90.81\%(\pm1.80\%)$ & $95.52\%(\pm0.18\%)$ & $90.57\%(\pm2.46\%)$ & $93.47\%(\pm0.10\%)$ & $89.09\%(\pm1.29\%)$  \\
& Entropy                  & $81.45\%(\pm0.22\%)$ & $79.21\%(\pm3.17\%)$ & $80.74\%(\pm0.31\%)$ & $80.84\%(\pm2.35\%)$ & $83.52\%(\pm0.34\%)$ & $83.80\%(\pm2.91\%)$  & $82.83\%(\pm0.18\%)$ & $82.81\%(\pm1.44\%)$  \\
& LMS $\cup$ MCDO  &                      &                      & $97.25\%(\pm0.09\%)$ & $\mathbf{91.17\%}(\pm1.75\%)$ &                      &                    & &   \\
\hline
\multirow{4}{*}{\rotatebox{90}{AUPRC}} & LMS                      & $91.29\%(\pm0.23\%)$ & $\mathbf{73.47\%}(\pm2.52\%)$ & $90.98\%(\pm0.25\%)$ & $74.35\%(\pm2.69\%)$ & $86.09\%(\pm0.31\%)$ & $\mathbf{64.54\%}(\pm5.38\%)$ & $70.82\%(\pm0.54\%)$ & $\mathbf{50.25\%}(\pm2.97\%)$  \\
& LMS w/o features & $87.73\%(\pm0.20\%)$ & $72.00\%(\pm2.67\%)$ & $87.34\%(\pm0.27\%)$ & $73.70\%(\pm2.91\%)$ & $81.11\%(\pm0.51\%)$ & $64.29\%(\pm5.38\%)$ & $65.49\%(\pm0.42\%)$ & $48.26\%(\pm3.31\%)$  \\
& Entropy                  & $49.16\%(\pm0.59\%)$ & $48.48\%(\pm5.75\%)$ & $48.16\%(\pm0.61\%)$ & $48.24\%(\pm5.99\%)$ & $47.25\%(\pm0.48\%)$ & $47.91\%(\pm4.53\%)$ & $35.16\%(\pm0.33\%)$ & $35.29\%(\pm2.88\%)$ \\
& LMS $\cup$ MCDO  &                      &                      & $91.04\%(\pm0.04\%)$ & $\mathbf{74.39\%}(\pm2.62\%)$ &                      &                      & & \\
\hline
\multicolumn{10}{|c|}{Regression $\sIoU$} \\
\hline
\multirow{4}{*}{\rotatebox{90}{$R^2$}} & LMS                      & $79.34\%(\pm0.19\%)$ & $\mathbf{66.69\%}(\pm2.06\%)$ & $78.19\%(\pm0.16\%)$ & $66.08\%(\pm2.23\%)$ & $74.04\%(\pm0.31\%)$ & $\mathbf{61.57\%}(\pm2.94\%)$ & $57.93\%(\pm0.37\%)$ & $\mathbf{48.84\%}(\pm1.81\%)$ \\ 
& LMS w/o features & $75.78\%(\pm0.18\%)$ & $64.91\%(\pm1.87\%)$ & $74.91\%(\pm0.18\%)$ & $65.31\%(\pm2.21\%)$ & $70.78\%(\pm0.30\%)$ & $61.51\%(\pm2.96\%)$ & $54.73\%(\pm0.19\%)$ & $47.62\%(\pm1.64\%)$ \\ 
& Entropy                  & $51.45\%(\pm0.37\%)$ & $50.88\%(\pm2.74\%)$ & $48.54\%(\pm0.57\%)$ & $48.21\%(\pm4.25\%)$ & $50.90\%(\pm0.45\%)$ & $50.30\%(\pm3.27\%)$ & $39.51\%(\pm0.28\%)$ & $39.33\%(\pm2.57\%)$  \\ 
& LMS $\cup$ MCDO  &                      &                      & $78.26\%(\pm0.14\%)$ & $\mathbf{66.13\%}(\pm2.27\%)$ &                      &                 & &     \\
\hline

\end{tabular}
}
\end{center}
\end{table*}

\subsection{NuScenes}

The nuScenes dataset \cite{caesar2020nuscenes} contains street scenes from two cities, Boston (US) and Singapore. It provides $700$ sequences for training and $150$ sequences for validation. Each sequence contains about $40$ samples which amounts to a total of $34$K key frames. The dataset has $16$ classes and is recorded and annotated with $2$ fps. The LiDAR sensor has $32$ channels and an angular resolution of \ang{0.33}. Every point cloud contains roughly $35$K points. For our experiments we used the pretrained Cylinder3D with the recommended data split. We did not test RangeNet++ and SalsaNext since the corresponding pretrained models are not available.

The image projection is of size $(w,h)=(1090, 32)$. Computing the connected components for all samples of the $150$ validation sequences yields approx.\ $1.5$M segments. Excluding all small segments containing less than $10$ points, i.e., $\SP< 10$, reduces that number to $0.34$M. Still, we retain $99 \%$ of the data in terms of points.
We performed $10$-fold cross validation where we always took $90 \% $ of the $150$ sequences, i.e., $135$ sequences, for training and the remaining $10 \%$, i.e., $15$ sequences for validation of the meta models. 
The results are presented in \cref{tab:summary}.
$89.85 \% $ of all segments have an $\sIoU >0$. With the meta classification we achieve an accuracy of $91.00 \%$, AUROC of $90.00 \%$ and AUPRC of $50.25 \%$, see `LMS' rows.
For the meta regression we achieve $R^2 = 49.19 \% $ for the validation data. The quality of predicting the $\sIoU$ is shown in \cref{fig:03_scatter_pred_true_iou}.

\subsection{Metric Selection}
So far, we have presented results based on all metrics from \cref{sec:method}, indicated by LMS in \cref{tab:summary}. 
In order to analyze the impact of the metrics to the performance, we repeated the experiments for multiple sets of metrics. 

\paragraph{Feature Measures}
First, we tested the performance of the meta classification and regression model without the feature measures, i.e., the metrics based on the point cloud input features, see row `LMS w/o features'. The performance in terms of ACC, AUROC, AUPRC and $R^2$ for all experiments are up to $2$ percentage points (pp.) lower compared to when incorporating feature measures.

\paragraph{Entropy}
Since the entropy is commonly used in uncertainty quantification, we tested all experiments with only using the mean entropy $\mu E$, see `Entropy' rows. The performance for the meta classification is up to $12$ pp.\ lower compared to LMS, for the meta regression $R^2$ decreases by up to $18$ pp. 

\paragraph{Bayesian Uncertainties}
The projection-based SalsaNext model follows a Bayesian approach as already mentioned in \cref{sec:introduction}: the LiDAR model provides a model (epistemic) and observation (aleatoric) uncertainty output for the point cloud's 2D image representation prediction, estimated by MC dropout (MCDO). To get these uncertainties we followed the procedure in \cite{cortinhal2020salsanext}. 
This ends up in epistemic $\mathit{epi}_l$ and aleatoric $\mathit{ale}_l$ uncertainty values for each pixel position $l$. We compute the same aggregated measures as for the measures $M \in \mathcal{M}$. Adding these new metrics to the previous metrics LMS is refereed to as LMS $\cup$ MCDO. The additional Bayesian uncertainties do not improve the meta classification and regression performance significantly, see \cref{tab:summary}. We have not tested SalsaNext on nuScenes since the pretrained model is not available. For comparability of results, we only used publicly available pretrained models.

\begin{table*}[t]
\begin{center}
\caption{Metric selection using a greedy method that in each step adds one metric that maximizes the meta classification / regression performance in terms of ACC / $R^2$ in $\%$. All results, SemanticKITTI (top) and nuScenes (bottom) are calculated on the dataset's metrics’ validation set.} \label{tab:metric-selection}
\scalebox{\tabscale}{

\begin{tabular}{|l|l|c|c|c|c|c|c|c|c|c|c|c|c|c|c|c||c|}
\hline
\multirow{16}{*}{\rotatebox{90}{SemanticKITTI}} & number of metrics & $1$ & $2$ & $3$ & $4$ & $5$ & $6$ & $7$ & $8$ & $9$ & $10$ & $11$ & $12$ & $13$ & $14$ & $15$ & $124$ \\
\cline{2-18}
 & \multicolumn{17}{|c|}{RangeNet++} \\
\cline{2-18}
& ACC & $81.13$ & $82.56$ & $83.35$ & $83.69$ & $83.96$ & $84.22$ & $84.40$ & $84.62$ & $84.78$ & $84.91$ & $85.01$ & $85.09$ & $85.09$ & $85.18$ & $85.25$ & $85.50$ \\
& Added & $\mu V$ & $P_{16}$ & $\upsilon \bar{F}^z_{in}$ & $P_{14}$ & $P_{13}$ & $\upsilon \bar{F}^y$ & $P_{15}$ & $\mu F^r_{bd}$ & $\mu F^i$ & $\upsilon F^z$ & $N_1$ & $P_{11}$ & $\upsilon F^i$ & $\mu D_{bd}$ & $P_1$ & all \\ 
 \cline{2-18}
& $R^2$ & $56.74$ & $58.20$ & $59.01$ & $59.87$ & $60.72$ & $61.57$ & $62.07$ & $62.69$ & $63.24$ & $63.70$ & $64.06$ & $64.43$ & $64.72$ & $64.91$ & $65.15$ & $66.69$ \\ 
& Added & $\mu V$ & $P_{14}$ & $P_{16}$ & $P_{17}$ & $P_{15}$ & $P_9$ & $\upsilon F^y_{bd}$ & $\upsilon \bar{F}^z$ & $\mu F^i$ & $P_{13}$ & $P_3$ & $\upsilon F^x$ & $\mu F^r$ & $N_{18}$ & $\mathit{SP}$ & all \\ 
\cline{2-18}

& \multicolumn{17}{|c|}{SalsaNext} \\
\cline{2-18}
& ACC & $82.05$ & $83.49$ & $84.10$ & $84.70$ & $84.93$ & $85.16$ & $85.35$ & $85.52$ & $85.65$ & $85.76$ & $85.86$ & $85.96$ & $85.96$ & $86.03$ & $86.11$ & $86.26$ \\
& Added & $\mu V$ & $P_{16}$ & $P_{15}$ & $P_{14}$ & $\upsilon F^y_{bd}$ & $P_{13}$ & $\mu \bar{D}$ & $\upsilon F^z$ & $N_1$ & $P_{12}$ & $\mu \bar{V}^z_{in}$ & $P_{10}$ & $P_{11}$ & $\upsilon F^i_{bd}$ & $\upsilon F^x$ & all \\ 
 \cline{2-18}
& $R^2$ & $56.18$ & $58.56$ & $59.59$ & $60.71$ & $61.67$ & $62.35$ & $62.87$ & $63.36$ & $63.77$ & $64.11$ & $64.40$ & $64.62$ & $64.86$ & $65.02$ & $65.21$ & $66.08$ \\ 
& Added & $\mu D$ & $P_{14}$ & $P_{17}$ & $P_{15}$ & $P_{13}$ & $\upsilon F^r$ & $\upsilon F^z$ & $P_{12}$ & $P_1$ & $P_9$ & $P_2$ & $\mu \bar{V}^i$ & $\upsilon F^y$ & $N_{10}$ & $\mu \bar{V}^z$ & all \\ 
\cline{2-18}

& \multicolumn{17}{|c|}{Cylinder3D} \\
\cline{2-18}
& ACC & $86.13$ & $86.76$ & $87.19$ & $87.41$ & $87.60$ & $87.75$ & $87.84$ & $87.93$ & $88.02$ & $88.08$ & $88.17$ & $88.26$ & $88.26$ & $88.29$ & $88.33$ & $88.37$ \\
& Added & $\mu D$ & $P_{15}$ & $\upsilon \bar{F}^r$ & $P_{17}$ & $P_{11}$ & $\upsilon D$ & $P_{13}$ & $P_{10}$ & $\upsilon \bar{F}^z$ & $N_1$ & $\mathit{SP}$ & $\mu F^r_{bd}$ & $N_{15}$ & $P_{14}$ & $\mu F^i$ & all \\ 
\cline{2-18}
& $R^2$ & $53.10$ & $54.54$ & $55.71$ & $56.76$ & $57.64$ & $58.36$ & $58.90$ & $59.31$ & $59.68$ & $60.03$ & $60.32$ & $60.66$ & $60.95$ & $61.12$ & $61.33$ & $61.57$ \\ 
& Added & $P_{14}$ & $\mu E_{bd}$ & $P_{17}$ & $P_{15}$ & $\mathit{SP}$ & $\upsilon \bar{F}^x$ & $\upsilon F^i$ & $P_6$ & $\upsilon F^y$ & $N_9$ & $P_{12}$ & $\upsilon D$ & $P_{13}$ & $P_5$ & $S$ & all \\ 
\hline
\multicolumn{18}{c}{} \\
\hline
\multirow{6}{*}{\rotatebox{90}{nuScenes}} & number of metrics & $1$ & $2$ & $3$ & $4$ & $5$ & $6$ & $7$ & $8$ & $9$ & $10$ & $11$ & $12$ & $13$ & $14$ & $15$ & $118$ \\
\cline{2-18}
& \multicolumn{17}{|c|}{Cylinder3D} \\
\cline{2-18}
& ACC & $90.06$ & $90.27$ & $90.44$ & $90.53$ & $90.61$ & $90.67$ & $90.73$ & $90.8$ & $90.85$ & $90.87$ & $90.89$ & $90.91$ & $90.91$ & $90.93$ & $90.95$ & $91.00$ \\
& Added & $\mu D$ & $\mu \bar{V}^r$ & $P_9$ & $P_{16}$ & $\mu F^x_{bd}$ & $\upsilon F^r$ & $P_{12}$ & $P_4$ & $\upsilon \bar{E}$ & $\mu \bar{V}^i$ & $N_{11}$ & $P_{15}$ & $\mu F^z_{bd}$ & $P_3$ & $\upsilon F^z_{bd}$ & all \\ 
\cline{2-18}
& $R^2$ & $40.88$ & $43.25$ & $44.27$ & $45.14$ & $45.79$ & $46.32$ & $46.80$ & $47.16$ & $47.4$ & $47.62$ & $47.79$ & $48.02$ & $48.1$ & $48.24$ & $48.33$ & $48.84$ \\ 
& Added & $\mu V_{bd}$ & $P_{16}$ & $\upsilon \bar{F}^r$ & $P_3$ & $P_{13}$ & $\mu F^r$ & $P_4$ & $\mu F^z$ & $\mu F^y_{bd}$ & $\mu F^i_{bd}$ & $N_{14}$ & $N_{10}$ & $\upsilon F^z$ & $P_{10}$ & $P_1$ & all \\ 
\hline

\end{tabular}
}
\end{center}
\end{table*}

\begin{figure}
    \centering
    \centerline{\includegraphics[width=\figwidth \textwidth]{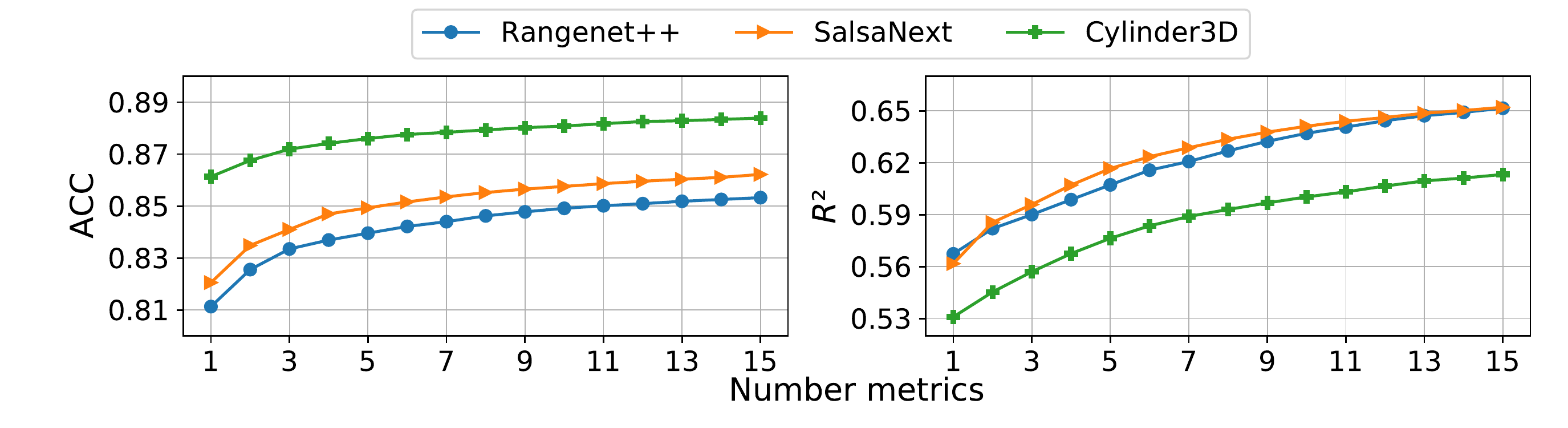}}
    \caption{Performance of the meta classification (left) and the meta regression (right) model on SemanticKITTI depending of the number of metrics, which are selected by the greedy approach.}
    \label{fig:04_metric_selection}
\end{figure}

\paragraph{Greedy Heuristic}
Inspired by fordward-stepwise selection for linear regression, we
analyze different subsets of metrics by performing a greedy heuristic: we start with an empty set of metrics and iteratively add a single metric that maximally improves the performance -- ACC for the false positive detection and $R^2$ for the prediction quality estimation. We performed this greedy heuristic for both, meta classification and meta regression. The results in terms of ACC and $R^2$ are shown in \cref{fig:04_metric_selection} (only for SemanticKITTI) and in \cref{tab:metric-selection}.
For the meta classification, we observe a comparatively big accuracy gain during adding the first $5$ metrics, then the accuracy increases rather moderately. 
For the meta regression, this performance gain in terms of $R^2$ spreads wider across the first 10 iterations, before the improvement per iteration becomes moderate.
Furthermore the results show that a small subset of metrics is sufficient for good models. We achieve nearly the same performance for both tasks with $15$ metrics selected by the greedy heuristic compared to when using all metrics (LMS). 
Considering \cref{tab:metric-selection}, the mean variation ratio $\mu V$ and the mean probability difference $\mu D$ in most cases constitute the initial choices.
Furthermore, the mean probabilities $P_i, \ i \in \mathcal{C}$, are also frequently subject to early incorporation.

\subsection{Confidence Calibration}

The false positive detection is based on a meta classification model, which classifies whether the predicted $\sIoU$ is equal or greater than $0$. In order to demonstrate the reliability of the classification model, we show that the confidences are well calibrated. 
Confidence scores are called \emph{calibrated}, if the confidence is representative for the probability of correct classification, cf.\ \cite{guo2017calibration}. 

The meta classification model estimates for each predicted segment the probability of being false positive, i.e., $\sIoU = 0$. We group the probabilities for all meta classified segments of the validation data into $10$ interval bins $( 0.0, 0.1 ], \ (0.1, 0.2], \ \ldots, \ (0.9, 1.0]$. The accuracy of a bin is the relative amount of true predictions; the confidence of a bin is the mean of its probabilities. The closer the accuracy and the confidence are to each other, the more reliable is the corresponding classification model. This is visualized in a so-called reliability diagram. For the evaluation of calibration, we define the maximum calibration errors (MCE) as the maximum absolute difference between the accuracy and the confidence over all bins and the expected calibration errors (ECE) as a weighted average of the bins' difference between accuracy and confidence, where the weights are proportional to the number of elements per bin. Further details are given in \cite{guo2017calibration}.

\begin{figure}
    \centering
    \centerline{\includegraphics[width=\figwidth \textwidth]{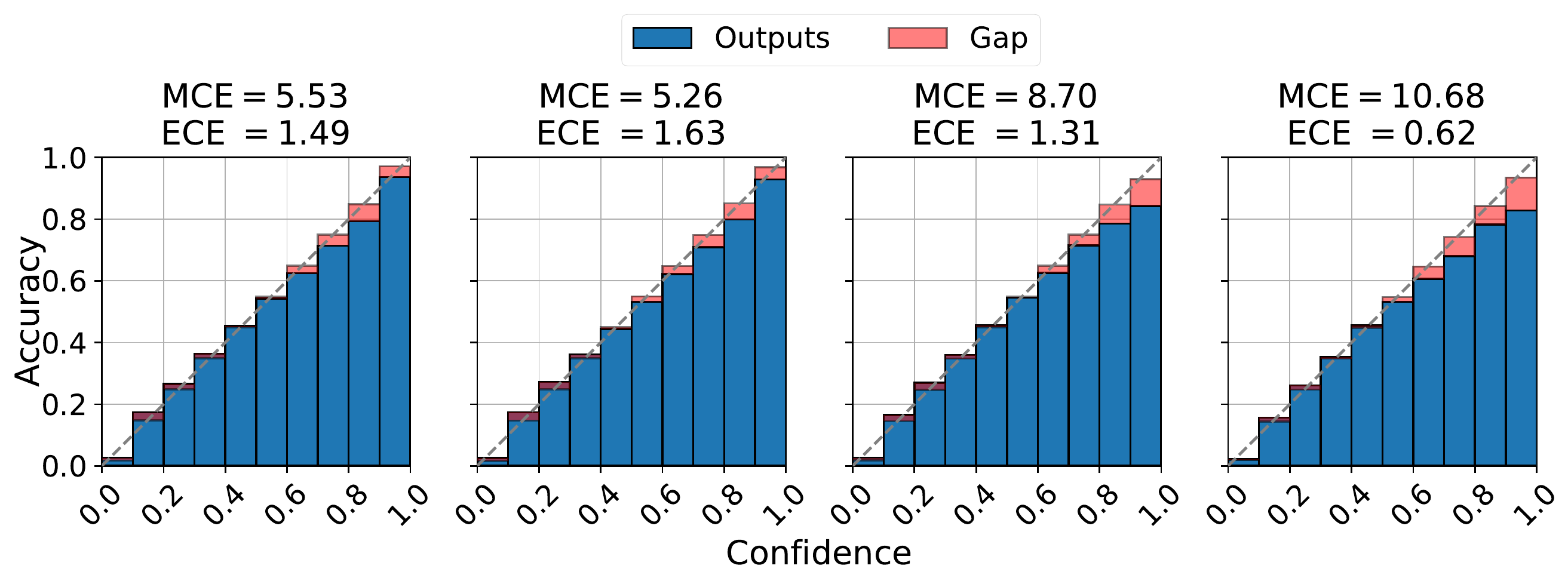}}
    \caption{Reliability diagrams with MCE and ECE for the meta classification model: RangeNet++, SalsaNext, Cylinder3D on SemanticKITTI as well as Cylinder3D on nuScenes, from left to right.}
    \label{fig:05_calib_plots}
\end{figure}

The reliability diagrams and the MCE as well as the ECE for all previously discussed meta classification models are shown in \cref{fig:05_calib_plots}. 
The smaller the gaps, i.e., the closer the outputs are to the diagonal, the more reliable and well calibrated is the model.
The MCE and ECE are between $5.26$ -- $10.68$ and $0.62$ -- $1.63$, respectively.
The results indicate well calibrated and reliable meta classification models.

\section{Conclusion} \label{sec:conclusion}

In this work we presented our method LidarMetaSeg for segmentwise false positive detection and prediction quality estimation of LiDAR point cloud segmentation.
We have shown that the more of our hand-crafted aggregated metrics we incorporate, the better the results get. This holds for all considered evaluation metrics -- ACC, AUROC, AUPRC and $R^2$. Furthermore, the results show that adding Bayesian uncertainties (epistemic and aleatoric ones approximated by MC dropout) on top of our dispersion measures based on the softmax probabilities neither improves meta classification nor meta regression performance. We have demonstrated the effectiveness of the method on street scene scenarios and are positive that this method can be adapted to other LiDAR segmentation tasks and applications, e.g.\ indoor segmentation or panoptic segmentation.
 
\bibliography{bib}
\bibliographystyle{unsrt}

\end{document}